\documentclass{article}

\usepackage{xcolor}
\definecolor{cvprblue}{rgb}{0.21,0.49,0.74}
\usepackage[pagebackref,breaklinks,colorlinks,allcolors=cvprblue]{hyperref}
\PassOptionsToPackage{numbers, sort&compress}{natbib}

\usepackage[preprint]{neurips_2026}

\usepackage{enumitem}
\usepackage{array}
\usepackage[utf8]{inputenc} 
\usepackage[T1]{fontenc}    
\usepackage{url}            
\usepackage{booktabs}       
\usepackage{amsfonts}       
\usepackage{nicefrac}       
\usepackage{microtype}      
\usepackage{graphicx}         
\usepackage{xspace}
\usepackage{amsmath}
\usepackage{wrapfig}

\definecolor{pendingcolor}{HTML}{C0C0C0}
\definecolor{fixedcolor}{HTML}{585858}
\definecolor{targetcolor}{HTML}{FFA600}
\definecolor{contextcolor}{HTML}{008CFF}

\newcommand{\pending}[1]{\textbf{\textcolor{pendingcolor}{#1}}}
\newcommand{\fixed}[1]{\textbf{\textcolor{fixedcolor}{#1}}}
\newcommand{\target}[1]{\textbf{\textcolor{targetcolor}{#1}}}
\newcommand{\context}[1]{\textbf{\textcolor{contextcolor}{#1}}}

\newcommand{\methodname}{\textsc{MilliVid}\xspace}
\newcommand{\loopcraft}{\textsc{Loopcraft}\xspace}

\makeatletter
\renewcommand{\paragraph}[1]{\par\textbf{#1}\space}
\makeatother

\title{\methodname: Hierarchical Latents for\\Long-Range Consistency in Video Generation}

\makeatletter
\def\@bottomtitlebar{%
  \vskip 0.29in \vskip -\parskip
  \hrule height 1\p@
  \vskip 0in}   
\makeatother

\makeatletter
\author{%
  Ishaan Preetam Chandratreya\thanks{Equal contribution. Order decided by coin flip.}\\
  {\normalsize Massachusetts Institute of Technology}\\
  {\normalsize\texttt{ishaanpc@mit.edu}}
  \vspace{-10pt}
  \And
  David Charatan\footnotemark[1]\\
  {\normalsize Massachusetts Institute of Technology}\\
  {\normalsize\texttt{charatan@mit.edu}}
  \vspace{-10pt}
  \AND
  \makebox[\textwidth][c]{\normalfont%
    \begin{tabular}[t]{c}\rule{\z@}{14\p@}\ignorespaces
      \textbf{Basile Van Hoorick}\\
      {\normalsize Toyota Research Institute}\\
      {\normalsize\texttt{basile.vanhoorick@tri.global}}
    \end{tabular}%
    \hspace{2em}%
    \begin{tabular}[t]{c}\rule{\z@}{14\p@}\ignorespaces
      \textbf{Sergey Zakharov}\\
      {\normalsize Toyota Research Institute}\\
      {\normalsize\texttt{sergey.zakharov@tri.global}}
    \end{tabular}%
    \hspace{2em}%
    \begin{tabular}[t]{c}\rule{\z@}{14\p@}\ignorespaces
      \textbf{Vitor Guizilini}\\
      {\normalsize Toyota Research Institute}\\
      {\normalsize\texttt{vitor.guizilini@tri.global}}
    \end{tabular}%
  }
  \vspace{-10pt}
  \AND
  Phillip Isola\\
  {\normalsize Massachusetts Institute of Technology}\\
  {\normalsize\texttt{phillipi@mit.edu}}
  \And
  Vincent Sitzmann\\
  {\normalsize Massachusetts Institute of Technology}\\
  {\normalsize\texttt{sitzmann@mit.edu}}
}
\makeatother

\begin{document}

\maketitle
\vspace{-16pt}


\begin{abstract}
Video generative models have become increasingly powerful, but long-range consistency remains challenging to achieve because even a few dozen frames require impractically long transformer sequence lengths.
We show that this issue can be mitigated by generating video using coarse-to-fine rollout within a multi-scale token space.
Our approach is simple: first, we pre-train an autoencoder that compresses each frame into a hierarchy of tokens, with levels ranging from the typical latent resolution to only a handful of tokens per frame.
The coarsest levels capture the most consequential information---such as scene layout and semantics---while finer levels add high-frequency appearance and texture.
Then, we train a video diffusion model to generate these tokens using coarse-to-fine rollout.
By carefully controlling the level of detail at which frames are generated and used as context during each rollout step, we are able to preserve long-range consistency in geometry and object permanence while spending less compute on the long-range consistency of less perceptually relevant details.
We validate this approach using a custom dataset of long Minecraft videos, where it produces substantially more consistent rollouts compared to existing baselines. Project page: \href{https://davidcharatan.com/millivid/}{\nolinkurl{davidcharatan.com/millivid}}.
\end{abstract}
\vspace{-12pt}
\section{Introduction}
Video generative modeling has advanced rapidly in realism, scale, and generality, with growing relevance to applications such as computer graphics, robotics, and world modeling~\cite{hafner2025dreamerv4,du2023learninguniversalpolicies,wang2025thisandthat,ye2026worldactionmodelszeroshot,chen2025largevideoplanner,po2026multigen}.
Yet long-range consistency---generating long videos that stay coherent from start to finish---remains an ongoing challenge.
Current models are largely designed for autoregressive rollout, generating long videos chunk by chunk, with the most recent chunk serving as context for the next.
This produces long videos, but sacrifices consistency: as chunks exit the context, their content is forgotten.

A straightforward solution to this problem is to increase the context length.
However, this quickly becomes prohibitive, as compute costs scale quadratically with sequence length in transformers and each additional video frame adds several hundred tokens.
A more efficient approach, exemplified by FramePack~\cite{zhang2025framepack}, is to allocate fewer tokens to distant context frames.
This strategy rests on the observation that different visual information needs to persist over different temporal horizons.
Coarse structure, such as the layout of a scene, must be faithfully preserved over long time spans, as any inconsistency is immediately noticeable.
Fine structure, such as exact texture patterns, can safely be forgotten, since long-range inconsistency within minor details is less perceptible.
However, we find that this insight alone does not tell the whole story---surprisingly, FramePack largely fails to recall content that leaves the camera frame, even if that content remains in its compressed context window.

We therefore present two further insights that enable consistent long-context video generation.
First, where prior work on token-efficient image and video generation has often compressed by simply downsampling or patchifying in either pixel or latent space, we find that training a hierarchical tokenizer yields compression that better preserves relevant detail. 
Second, we hypothesize that a model trained for predicting only short future time horizons learns to only consider the most recent context, mirroring insights for training long-horizon policies in robot behavior cloning ~\cite{villasevil2025learning,chi2023diffusion,zhao2023learning}. 
We hence force the model to make predictions \emph{far into the future}.

Based on these insights, we present \methodname, a diffusion-based video generative model designed to maximize long-range consistency under a fixed sequence length.
Our approach has two components.
First, inspired by flexible image tokenizers~\cite{bachmann2025flextok,alit}, we train an autoencoder with a hierarchical latent space, in which each level represents an image using a specific number of tokens.
In this latent space, coarser levels capture global structure, while finer levels add detailed appearance and texture.
Second, we train a video diffusion model for coarse-to-fine generation in this latent space.
Generation starts at the coarsest level, where the small number of tokens per frame allows the model to cover many frames, and then progressively refines toward finer scales.
Transformer weights are shared across all scales; the same transformer with a fixed sequence length performs generation at every scale.

We evaluate our method on long videos of Minecraft gameplay, which we find to be particularly suitable for measuring long-range consistency.
Compared with prior work on long-context video generation, our method produces substantially more consistent long rollouts, successfully recalling details and scene structure that baselines forget, without relying on retrieval or explicit 3D maps.
Our contributions are as follows:
\begin{itemize}[leftmargin=16pt, noitemsep]
\vspace{-3pt}
    \item We show that hierarchical autoencoding can be combined with a novel coarse-to-fine rollout strategy to achieve long-range consistency in video generation.
    \item We demonstrate that under a common sequence length constraint, our proposed method produces significantly better consistency than FramePack, the state-of-the-art baseline, as well as typical autoregressive rollout, without sacrificing per-frame quality.
\vspace{-6pt}
\end{itemize}

\section{Related Work}

\paragraph{Retrieval-augmented video generation} seeks to extend the effective temporal context of a video generative model by retrieving a small set of frames or chunks from the distant past and appending them to the current context window~\cite{cai2025mixture,yu2025context,xiao2025worldmem, wang2026matrixgame30realtimestreaming, song2025generative}. 
Our approach fundamentally differs in that we do not perform retrieval and instead rely only on a coarse-to-fine hierarchy to extend the temporal context of the transformer. Our method is orthogonal and could be combined with a retrieval mechanism for past frames that fall outside the temporal context of the coarsest level of our hierarchy. 

\paragraph{3D Memory} A particularly effective way to store information about past generations is to leverage 3D geometry.
\citet{garcin2026beyond} use unprojection and projection into a 3D voxel grid for persistent 3D scene memory.
\citet{huang2025memory} build an incremental 3D point cloud and use camera pose to retrieve past frames most relevant to the current target frames. We pursue a conceptually different direction that assumes neither camera poses nor an explicit 3D world.

\paragraph{Flexible-length tokenization} learns representations whose token length can vary with the desired compression level. In images, variable-length tokenization has been shown to naturally induce a coarse-to-fine ordering that can be leveraged for coarse-to-fine image generation~\cite{bachmann2025flextok,wen2025principal,liu2025detailflow,elastictok}. SceneTok~\cite{asim2026scenetok} demonstrates that flexible-length tokenization can similarly accelerate 3D scene generation for novel view synthesis. Flexible-length \emph{spatiotemporal} tokenization has been explored for video to facilitate efficient generation~\cite{atanov2026videoflextok,xiong2026evatok,elastictok}, but does not propose methods to extend long-range video consistency. By contrast, we use a \emph{per-frame} tokenizer in conjunction with a latent diffusion model to demonstrate long-range consistent video generation.
Our frame tokenizer can be considered a multi-scale variant of ElasticTok~\cite{elastictok}, and is related to other recent works in adaptive tokenization of images and video~\cite{flextok, matryoshka, alit, duggal2026singlepass}.

\paragraph{Multi-scale generation} is a classical way to reduce the cost of generative modeling by generating coarse structure first and refining detail later. In images, this idea appears in hierarchical latent models such as NVAE~\cite{vahdat2020nvae}, cascaded diffusion models~\cite{ho2022cascaded,mukhopadhyay2026scale}, and more recent coarse-to-fine generators such as Edify Image and VAR~\cite{atzmon2024edify,tian2024visual}. In video, Imagen Video and I2VGen-XL use cascaded diffusion pipelines~\cite{ho2022imagen,2023i2vgenxl}, while Pyramidal Flow Matching organizes generation across pyramid stages in a unified model~\cite{jin2024pyramidal}. 
TECO~\cite{yan2023temporallyconsistenttransformersvideo} achieves a long temporal context using aggressive spatial downsampling of latents before temporal attention.
FAR~\cite{gu2025long} proposes separate patchification schemes for far-away and recent frames.
FramePack~\cite{zhang2025framepack} is the most related and recent approach to multi-scale video modeling for long-range consistency. It proposes to downsample past frames' latents, enabling a longer past context window, but generates full-resolution target latents. Our method is likewise coarse-to-fine, but differs from prior work in two key ways: we learn the scale space itself, and we explicitly allocate a transformer's fixed token budget across scales, generating frames coarse-to-fine, to maximize temporal context and long-range consistency.

\section{Method}
In this section, we present \methodname, a two-stage algorithm for generating videos with long-term consistency.
First, in Section \ref{sec:adaptive}, we introduce an autoencoder that encodes frames into a hierarchical latent space, where each level contains a different number of tokens.
Next, in Section \ref{sec:diffusion}, we introduce a video generative model that is trained within this hierarchical latent space.
Given a fixed transformer sequence length, this model can flexibly alternate between generating many highly compressed frames and fewer highly detailed frames.
We harness this ability by designing a sampling algorithm that leverages both regimes, first generating long, highly compressed sequences, and then gradually upsampling them to generate fine details.

\subsection{Hierarchical Autoencoding}
\label{sec:adaptive}
\begin{figure*}[t!]
\centering
\makebox[\textwidth][l]{\includegraphics[width=403pt]{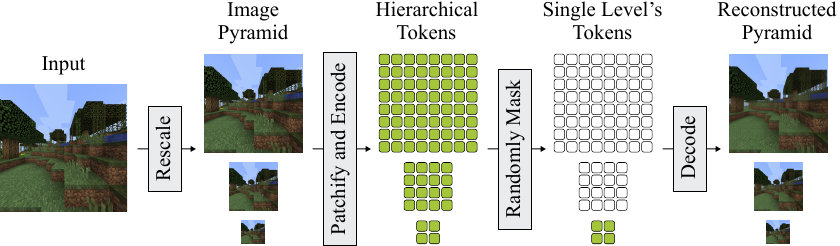}}
\caption{
Our hierarchical autoencoder consists of a multi-resolution encoder-decoder pair.
The encoder receives a multi-resolution image pyramid as input.
It patchifies the pyramid's images using a fixed kernel size, yielding fewer tokens for coarser levels, then feeds the tokens through a transformer.
During training, we zero out all but one randomly chosen level.
The decoder, also a transformer, must reconstruct the entire resolution cascade based on the remaining tokens.
The highest-resolution reconstruction is supervised using MSE and LPIPS, while the others are supervised only using MSE.
We show three levels and a small number of tokens here for clarity; in practice, we use more.
}
\vspace{-6pt}
\label{fig:autoencoder}
\end{figure*}
Our hierarchical autoencoder is designed to trade off between visual fidelity and token count.
As shown in Figure~\ref{fig:autoencoder}, it encodes a single video frame into a hierarchy of latent representations with a fixed number of levels.
At level $\ell$, the latent representation uses the following number of tokens:
$$
\textbf{per-frame token count $N_\ell$ at level $\ell$} \qquad N_\ell = \frac{H}{2^\ell} \times \frac{W}{2^\ell}
$$
The finest representation is at $\ell = 0$, where the frame is encoded into a grid of $H \times W$ tokens.
Each coarser level halves the previous level's resolution along both the height ($H$) and width ($W$) dimensions.
Coarser levels are more compressed, while finer levels yield higher-quality reconstructions.
Each level can be decoded on its own without access to the other levels' tokens.

Our encoder is a transformer that simultaneously outputs the full latent hierarchy.
To encode an image, we first transform it into a multi-resolution pyramid containing one image per hierarchy level.
Within this pyramid, the highest-resolution image matches the input frame's resolution, while the subsequent images each halve the previous image's height and width.
We patchify the pyramid's images using a fixed kernel size (with shared weights) across levels to produce $\frac{H}{2^\ell} \times \frac{W}{2^\ell}$ tokens per level; add positional encodings for each token's row, column, and level index; and then feed all of the tokens through the encoder.
This produces the complete latent hierarchy---the first $H \times W$ output tokens correspond to $\ell = 0$, the next $\frac{H}{2} \times \frac{W}{2}$ correspond to $\ell = 1$, and so on. 

Our decoder is a transformer that reconstructs the input frame from a single level's latent tokens.
It returns the input frame's reconstruction as the highest-resolution image in a multi-resolution pyramid that mirrors the encoder's input.
To decode a particular level's tokens, we take the encoder's output and zero out the other levels' tokens.
We then add positional encodings for row, column, and level; feed the tokens through the decoder; and unpatchify the output, mirroring the encoder's patchification.

To train the autoencoder, we randomly sample a level to decode at, then supervise on the resulting reconstruction.
Regardless of which level was sampled, we supervise on the entire output pyramid using MSE and also supervise on the highest-resolution image using LPIPS~\cite{zhang2018unreasonable}.
%
The lower-resolution output images serve to accelerate and stabilize convergence; they are discarded during inference.

\subsection{Coarse-to-Fine Video Generation}
\label{sec:diffusion}
\begin{figure*}[t!]
\centering

\noindent\hspace*{-20pt}%
\includegraphics[width=\dimexpr\linewidth+20pt\relax]{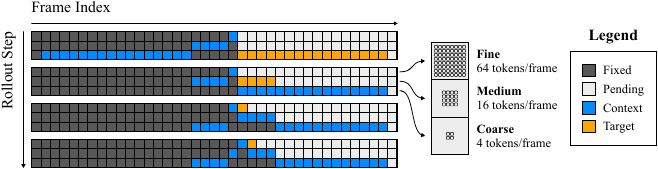}%
\caption{
To generate videos, we use coarse-to-fine rollout.
During each rollout step, the model sees a mixture of \context{context} and \target{target} tokens across different hierarchy levels; it does not see the \fixed{fixed} (already generated) or \pending{pending} (not yet generated) tokens.
Our model's first four rollout steps are shown on the left.
Each step is represented as a 3-row grid where the rows represent hierarchy levels and each column represents a single frame.
Over the first three steps, the model generates a long sequence of coarse frames, a medium-length sequence of medium frames, and a single fine frame.
The full rollout sequence is shown on our accompanying project page and in Figure~\ref{fig:diffusion_sampling_long} in the appendix.
We show three levels here for clarity; in practice, we use more.
}
\label{fig:diffusion_training}
\vspace{-6pt}
\end{figure*}
In this section, we describe our video generative model, which is a transformer-based latent diffusion model that operates within our autoencoder's hierarchical latent space.

The fundamental constraint that our model is designed around is a transformer's sequence length $S$.
Consider that in a typical video diffusion model, a video is represented as $F \times H \times W$ tokens.
For a 30-second, 20 fps video with $16 \times 16$ tokens per frame, this equals $600 \times 16 \times 16 = 153{,}600$ tokens.
In all but the most compute-rich settings, this exceeds feasible values of $S$, precluding us from feeding the entire video to the transformer at once.
Thus, we must design video models that operate under the constraint $S \ll F \times H \times W$ while still producing temporally coherent long videos.

Prior work has addressed this problem by using autoregressive rollout, which is subject to rapid forgetting.
Consider the case of $S = 1024$, which is enough for two context frames and two generated frames.
In this case, we can generate $F$ frames in two-frame chunks, but only have 0.1 seconds of context.
As a result, anything that even momentarily exits the frame will be forgotten.

\begin{wrapfigure}{r}{119pt}
    \centering    \includegraphics[width=119pt]{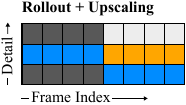}
    \vspace{-10pt}
\end{wrapfigure}

Our solution to this problem leverages our hierarchical autoencoder's ability to control the number of tokens allocated to each frame.
Broadly speaking, instead of rolling out at the full resolution, we roll out at the coarsest resolution---where many frames fit into $S$---then upscale the resulting frames.
The most obvious upscaling strategy, shown in the inset, conditions the \target{generated} frames on \context{context} consisting of the recent past and the next-coarser level.
Unfortunately, as we will show, this approach is doomed to generate inconsistencies in the resulting video.

To illustrate why this happens, consider a video in which the camera approaches a street sign whose text is currently too small to decipher.
When we play the video, the sign grows closer, and the text becomes legible.
If we could instead accurately super-resolve the video, we could read the text from far away.
In other words, uncertainty (about the sign's text) is resolved both with increased resolution and temporal rollout.
The issue with the aforementioned strategy is that it \emph{independently} performs super-resolution and temporal rollout.
If super-resolution resolves the sign's text as ``stop'' and temporal rollout resolves it as ``go,'' we are left with an awkward inconsistency.

Our method, shown in Figure~\ref{fig:diffusion_training}, solves this problem by carefully considering which tokens to use as context to avoid creating inconsistencies.
Like the aforementioned strategy, it alternates between generating many coarse frames and fewer fine frames at once.
However, critically, its context always includes both the highest-resolution most-recent frame \emph{and} future frames that have only been generated up to coarser levels, ensuring that situations like the stop-go discrepancy cannot occur.
Consult Figure~\ref{fig:diffusion_sampling_long} for an overview of the exact rollout procedure.

\paragraph{Training Procedure}
As shown in Figure~\ref{fig:diffusion_sampling_long}, our model's rollout consists of many similar but distinct steps.
At training time, we randomly sample from these steps and supervise the model on its ability to denoise the \target{generated} tokens conditioned on clean \context{context} tokens; the other tokens are not shown to the model.
An important property of our method is that each rollout step requires the same number of tokens (without padding).
As a result, different rollout steps can trivially be stacked into the same batch for efficient training.
We use a single model for all rollout steps, and positional encodings for each token's level, frame, row, and column let the model distinguish between steps.

\section{Results}
\begin{figure*}[t!]
\vspace{-4pt}
\centering
\hspace*{-0.2281in}%
\includegraphics{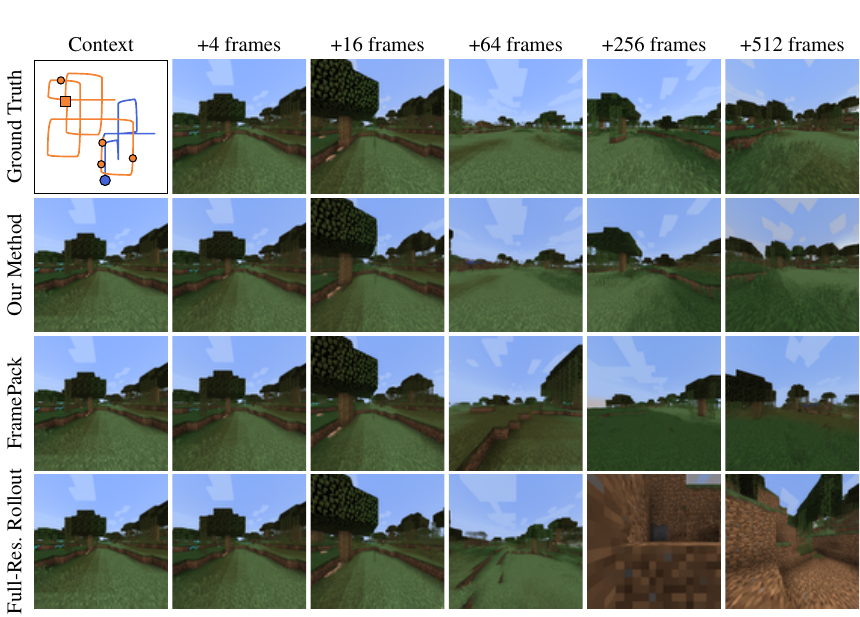}
\caption{
Unlike the baselines, our method reliably recalls the scene's structure, even when many frames have elapsed.
The top-left panel shows a top-down view of the trajectory the models follow.
Blue is context; orange is generated.
The points at which frames are shown are marked.
We encourage the reader to consult our project website for videos of similar comparisons.
}
\vspace{-6pt}
\label{fig:baselines_qualitative}
\end{figure*}
In this section, we validate our model's long-range consistency against FramePack~\cite{zhang2025framepack}, the most similar state-of-the-art approach to our method, as well as full-resolution autoregressive rollout.
\paragraph{Datasets}
A useful dataset for measuring long-range consistency must have four properties.
First, it must contain videos that are long---several hundred frames or more.
Second, it must provide enough videos to facilitate training a generative model.
Third, it must contain videos that can be used to measure memory, where the same content frequently exits and later re-enters the frame.
Finally, it must provide fine-grained conditioning signals (e.g., poses or actions) that can be used to steer video models towards previously-seen content.

Existing datasets fall short of these criteria.
For instance, video datasets like Kinetics~\cite{kay2017kinetics} do not provide appropriate fine-grained conditioning.
Posed video datasets like RealEstate10k~\cite{realestate10k}, ACID~\cite{acid_dataset}, and DL3DV-10k~\cite{ling2024dl3dv} contain relatively short videos or too few instances of content exiting and re-entering the frame.
Visual odometry datasets like KITTI~\cite{kitti_dataset} and TUM RGB-D~\cite{tum_rgbd_dataset} contain too few sequences for training.

We address this issue by generating \loopcraft, a dataset of 1024-frame videos of Minecraft gameplay.
It contains 200,000 videos at $256 \times 256$ resolution, with metadata for both action and pose conditioning.
Created using a dataset generation pipeline adapted from TECO's~\cite{yan2023temporallyconsistenttransformersvideo}, it uses the MineRL simulator~\cite{guss2019minerl} to collect trajectories of an agent running through the world and intermittently making random 90-degree turns.
The agent is slightly biased towards making pairs and quartets of turns, and as a result, it frequently returns to previously seen areas of the world.
This makes the \loopcraft dataset ideally suited to measuring long-range consistency.

\paragraph{Metrics}
\begin{table}[t]
\centering
\caption{
Consistency and quality metrics averaged over short (frames 1-64), medium (frames 65-256), and long horizons (frames 257-768).
We measure consistency using PSNR, LPIPS, SSIM, DINOv2 class token cosine similarity, and LightGlue matches.
Our LightGlue match metric counts the number of keypoint matches detected by LightGlue with confidence greater than 0.5. We measure quality using FID and FVD.
In each column, the best-performing method is highlighted in \textbf{bold}.
See Figure~\ref{fig:baselines_quantitative_extended} in the appendix for a plot showing a fine-grained view of the information presented in this table.
}
\label{tab:baselines}
\footnotesize
\setlength{\tabcolsep}{3pt}
\renewcommand{\arraystretch}{1.05}
\begin{tabular*}{\linewidth}{@{\extracolsep{\fill}}l|ccccccccc}
\toprule
& \multicolumn{3}{c|}{PSNR $\uparrow$}
& \multicolumn{3}{c|}{LPIPS $\downarrow$}
& \multicolumn{3}{c}{SSIM $\uparrow$} \\
Method
  & short & med. & long & short & med. & long & short & med. & long \\
\midrule
  Our Method        & \textbf{21.78} & \textbf{19.13} & \textbf{16.69} & \textbf{0.159} & \textbf{0.247} & \textbf{0.335} & \textbf{0.612} & \textbf{0.562} & \textbf{0.522} \\
  FramePack         & 19.17 & 14.08 & 11.98 & 0.238 & 0.437 & 0.533 & 0.607 & 0.491 & 0.434 \\
  Full-Res. Rollout & 17.58 & 11.95 & 11.02 & 0.290 & 0.538 & 0.630 & 0.583 & 0.439 & 0.407 \\
\bottomrule
\end{tabular*}
\vspace{4pt}
\begin{tabular*}{\linewidth}{@{\extracolsep{\fill}}l|cccccccccccc}
\toprule
& \multicolumn{3}{c|}{DINOv2 $\uparrow$}
& \multicolumn{3}{c|}{Matches $\uparrow$}
& \multicolumn{3}{c|}{FVD $\downarrow$}
& \multicolumn{3}{c}{FID $\downarrow$} \\
Method
  & short & med. & long & short & med. & long & short & med. & long & short & med. & long \\
\midrule
  Our Method        & \textbf{0.943} & \textbf{0.927} & \textbf{0.906} & \textbf{166.4} & \textbf{94.5} & \textbf{62.4} & \textbf{43.8} & \textbf{57.8} & \textbf{83.7} & \textbf{23.06} & \textbf{26.46} & \textbf{29.13} \\
  FramePack         & 0.920 & 0.854 & 0.803 & 138.7 & 22.5 & 7.0 & 64.9 & 145.0 & 286.2 & 24.92 & 39.89 & 57.72 \\
  Full-Res. Rollout & 0.906 & 0.824 & 0.754 & 144.6 & 12.7 & 5.0 & 78.1 & 211.6 & 607.1 & 24.69 & 41.06 & 76.96 \\
\bottomrule
\end{tabular*}
\vspace{-20pt}
\end{table}

\begin{figure*}[t!]
\centering
\hspace*{-0.2391in}%
\includegraphics{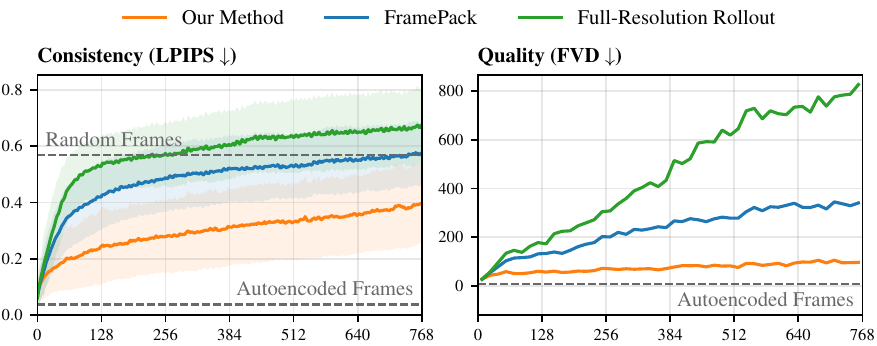}
\caption{
We separately measure consistency and quality on the \loopcraft dataset.
The X axis indicates the number of frames each model has generated since seeing ground-truth context.
We use LPIPS to measure consistency---i.e., how well generated videos match ground-truth videos following the same trajectories.
We use Fr\'echet Video Distance (FVD) as a measure of visual fidelity that is independent of consistency.
Our method matches the baselines on quality and clearly exceeds them on consistency.
Refer to Figure~\ref{fig:baselines_quantitative_extended} in the appendix for further plots of consistency and quality metrics.
}
\label{fig:baselines_quantitative}
\vspace{-6pt}
\end{figure*}
Our metrics are designed to independently measure two aspects of video generation: \underline{consistency} and \underline{quality}.
Consistency is a model's ability to accurately recall previously seen content; quality is the ability to generate high-quality frames during rollout, regardless of whether those frames are consistent or not.
The distinction between consistency and quality has previously been studied by~\cite{wang2025erroranalysesautoregressivevideo} and referred to as the drifting-forgetting tradeoff~\cite{zhang2025framepack}.

To measure consistency, we condition each video model on up to 256 frames (depending on its supported context length) of a 1024-frame ground-truth sequence, plus the ground-truth actions for all 1024 frames, then compare its 768-frame generated rollout against the ground truth.
Effectively, the question we ask is: if the model sees the world and then follows a prescribed path through it, can it reproduce the world's content correctly?
Our comparison uses image metrics---peak signal-to-noise ratio (PSNR), structural similarity index (SSIM)~\cite{wang2004image}, and perceptual similarity~\cite{zhang2018unreasonable}---in addition to DINOv2 class token cosine similarity~\cite{oquab2023dinov2} and LightGlue keypoint match count~\cite{lindenberger2023lightglue}.

We empirically find that measuring consistency can be challenging for two reasons.
First, it is difficult to guarantee that the content the models are asked to reproduce has been seen in the context.
To alleviate this issue, we generate a 1,000-video test set whose trajectories have high overlap between the context and the remaining trajectory (details in appendix \ref{subsec:test_set_generation}).
Second, even well-performing models can produce trajectories that drift slightly compared to the ground truth.
This manifests as slight misalignments in otherwise consistent scenes.
Several of our metrics (LPIPS, DINOv2, and keypoints) are robust to such shifts, and we empirically find that LPIPS is an especially good measure of consistency in the face of slight misalignments, as previously reported in~\cite{park2021nerfies}.

To measure quality, we use two metrics: Fr\'echet Inception Distance (FID) and Fr\'echet Video Distance (FVD).
We find that these metrics roughly correlate with perceived quality.

\paragraph{Models}
We compare our model to FramePack~\cite{zhang2025framepack} and full-resolution autoregressive rollout.
Since the baselines do not use hierarchical tokenization, we train them using our autoencoder's highest-resolution latent space.
All models share the same decoder, which decodes from this latent space to images.
All models are action-conditioned; the action space consists of a single ternary value that indicates whether the agent is turning left, turning right, or moving forward.
We describe the models below; see appendix ~\ref{subsec:training_details} for further training details.

\underline{\methodname (Our Method):}
We train our model with 4 latent levels.
In descending order of resolution, we train our model with 3, 12, 48, and 192 frames of context, which yields a total budget of 3840 tokens.
As a result, our model simultaneously denoises up to 192 frames (at the coarsest level) with up to 255 frames of context.

\underline{FramePack:}
In descending order of resolution, FramePack is trained with 3, 12, 48, and 192 frames of context.
While FramePack uses varying patchification of full-resolution latents rather than hierarchical latents, the number of tokens per frame of context matches our model exactly.
We train FramePack to denoise three high-resolution frames at once, which yields a token budget of exactly 3840 tokens.

\underline{Full-Resolution Autoregressive Rollout:}
This model denoises full-resolution target latents conditioned on full-resolution context latents.
We train this model with seven frames of context and seven denoised frames, for a token budget of 3584 tokens.

\paragraph{Results}
Our method clearly outperforms the baselines in terms of consistency and quality.
See Figure~\ref{fig:baselines_qualitative} for qualitative results, Table~\ref{tab:baselines} for quantitative results, and Figure~\ref{fig:baselines_quantitative} for a fine-grained view of how consistency changes as rollout progresses.
An expanded version of Figure~\ref{fig:baselines_quantitative} is available in the appendix as Figure~\ref{fig:baselines_quantitative_extended}.
We further urge the reader to consult our accompanying project page, which includes video results, to qualitatively judge the presented models.

\section{Analysis}
\label{sec:analysis}
We investigate our core design choices---hierarchical autoencoding and coarse-to-fine generation within a hierarchical latent space---against reasonable alternatives through three central questions.

\begin{figure*}[t!]
\vspace{-18pt}
\centering
\hspace*{-0.2281in}%
\includegraphics{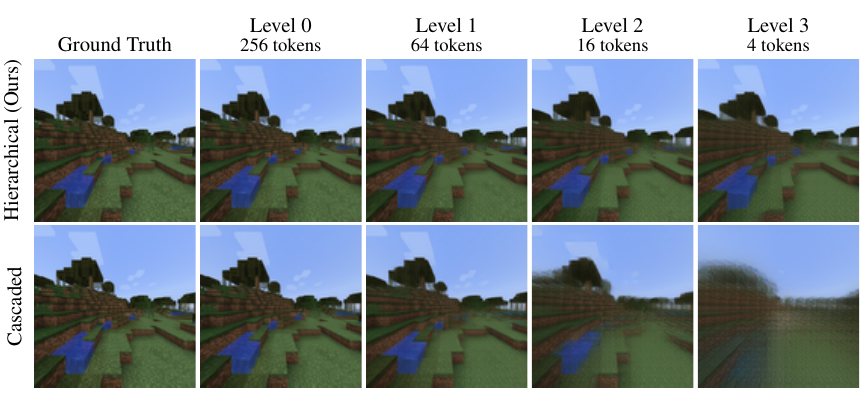}
\caption{
As our hierarchical autoencoder's token budget decreases, it discards fine-grained texture and geometry while retaining coarse scene structure (top row).
Compared to an autoencoder trained to decode mean-pooled, full-resolution latents (bottom row)---the kind of latents one would find in a cascaded diffusion model---our hierarchical autoencoder produces significantly better reconstructions.
} 
\label{fig:autoencoder_qualitative}
\end{figure*}

\paragraph{Q1: What information is contained within our hierarchical latent space?}
To answer this question, we visualize our hierarchical autoencoder's latent space.
See Figure~\ref{fig:autoencoder_qualitative} (top row) for an example of such visualization.
We find that our highest-resolution latent level faithfully recreates the input image, including both structure and texture, while the most compressed latent level retains coarse scene structure and forgets exact textures.
We highlight that this is significantly more desirable than simply producing a low-frequency (i.e., blurry) reproduction of the input image, since the most important scene structure remains clearly defined.

To highlight the benefits of our hierarchical latent space, we train a variant of our autoencoder in which the coarser levels are defined as downscaled (mean-pooled) versions of the finest level rather than being learned.
In this case, the decoder sees a resolution cascade, exactly as a cascaded diffusion model would.
Figure~\ref{fig:autoencoder_qualitative} (bottom row) shows the results of this experiment: the best the decoder can do is produce a blurry version of the input image, where most structure is lost.

\paragraph{Q2: How does a hierarchical model compare to an identically trained cascaded one?}
\begin{figure*}[t!]
\centering
\noindent\hspace*{-0.2391in}%
\includegraphics{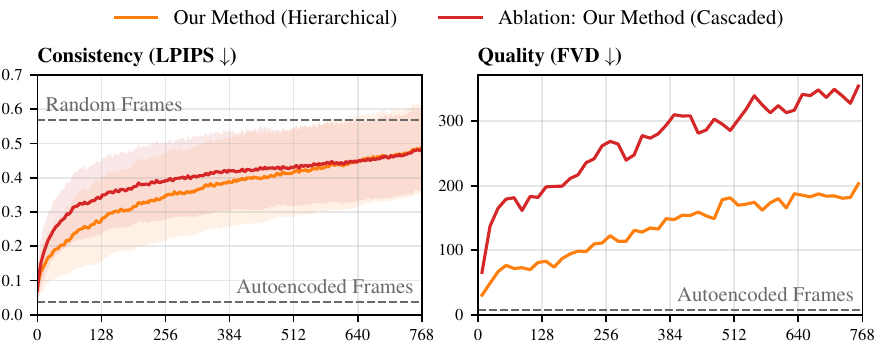}%
\caption{
A cascaded variant of our method, in which the model is trained to operate on downscaled (mean-pooled) versions of our highest-resolution hierarchy level instead of using the full latent hierarchy, performs worse on both consistency and quality.
For all ablations, we train models with sequence lengths of 1280, where 1/3 as many frames are seen at every level compared to the models in Figure~\ref{fig:baselines_quantitative}.
Note that the mean-pooled tokens the cascaded model sees are also visualized in Figure~\ref{fig:autoencoder_qualitative}.
}
\label{fig:ablation_cascaded}
\vspace{-6pt}
\end{figure*}
To answer this question, we train a cascaded variant of our method.
In this version, rather than predicting our highest-resolution latents from a series of coarser hierarchical latents, the model operates on a series of downscaled (mean-pooled) versions of the highest-resolution latents.
Figure~\ref{fig:ablation_cascaded} shows that this variant of our model is clearly worse, both in terms of consistency and quality.

\paragraph{Q3: Does FramePack benefit from hierarchical latents, and do models that can access many frames of context inherently give better consistency?}
\begin{figure*}[t!]
\centering
\noindent\hspace*{-0.2391in}%
\includegraphics{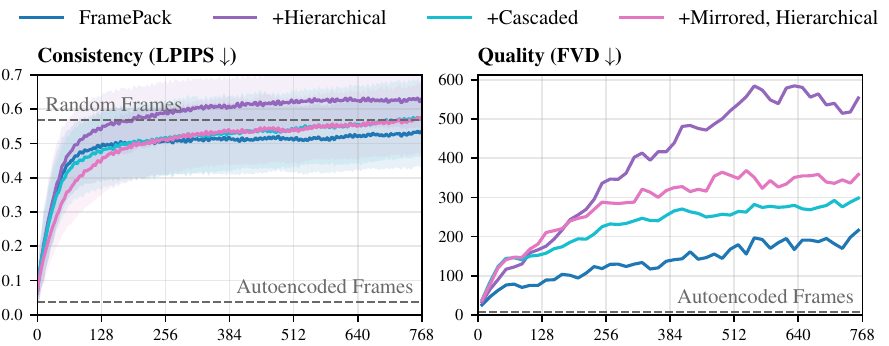}%
\caption{
We plot three variants of FramePack against the original FramePack model.
See Section~\ref{sec:analysis} (Q3) for a description of each model.
The mirrored variant performs better on consistency despite having a shorter context window; all perform worse on quality.
We find that the hierarchical variant's consistency becomes worse than random frames due to rollout instability.
\vspace{-12pt}
} 
\label{fig:framepack_quantitative}
\end{figure*}
We analyze three variants of FramePack: the original model, whose adaptivity stems from varying patchification kernel sizes; a variant trained using our hierarchical latent space; and a cascaded variant that uses the same cascaded latent space as in Figure~\ref{fig:ablation_cascaded}.
We also modify FramePack to allocate half of its capacity to the target, such that it not only has coarse context, but also coarse targets that exactly mirror the context's token layout.
This ``mirrored'' FramePack model predicts coarse, long-range targets at training time, but discards them at sampling time, where it behaves like a FramePack model with about half the usual context length.
See Figure~\ref{fig:framepack_quantitative} for the results of these experiments.
We find that the hierarchical and cascaded variants perform worse across both consistency and quality.
Surprisingly, despite having a shorter context window, the mirrored variant performs better than FramePack on consistency at certain time horizons.
We hypothesize that this is because, despite having a long context window, the original FramePack model only predicts a short distance into the future, where it can mostly rely on information from the most recent context frame.
By being asked to predict further into the future, the mirrored model may be forced to examine the distant past for information, improving its overall consistency.

\vspace{-4pt}
\section{Conclusion}
In this paper, we observe that reasoning over long visual sequences necessarily means forgetting \emph{something}: given a fixed sequence length, we have to decide on what past and present information to fill it with.
We suggest that in long video generation, we should compromise on the long-range consistency of fine detail, trading it off against an increased temporal horizon for coarse structure.
We hence learn a coarse-to-fine hierarchy by training an adaptive-length frame tokenizer, and then fill our token budget preferentially with many past coarse frames and only a few recent fine frames. 

Our study suggests that such multi-scale video generative modeling can yield dramatic gains in long-range consistency, recalling 3D scene structure over hundreds of frames where a conventional video generative model can only afford a handful of context frames.
Beyond its practical potential, we hope that our work will encourage future exploration of multi-scale video generative modeling.

\paragraph{Limitations and Future Work}
While our method fulfills its goal of achieving long-range consistency, it is not designed to be fine-tuned from standard video diffusion models.
As a result, fine-tuning a large, pre-trained model like WAN~\cite{wan2025wanopenadvancedlargescale} or Hunyuan Video~\cite{kong2024hunyuanvideo} is not straightforward.
We hope that future work will explore doing so---for example, one could distill these methods' original autoencoders to be hierarchical (e.g., by aligning a hierarchical autoencoder's finest latent space to the original latent space), then fine-tune the associated latent diffusion model to be hierarchical.
Additionally, compared to FramePack, our model requires about 33\% more rollout steps (see \ref{subsec:sampling_speed} in the appendix).
However, we believe this is a worthwhile tradeoff for the dramatically increased consistency our method offers.

\section{Acknowledgements}
We thank Andrew Song and Hannah Schlueter for their feedback during the process of writing and editing the paper. This work was supported by the Toyota Research Institute (TRI) University 3.0 (URP) program, the National Science Foundation under Grant No. 2211259, by the Intelligence Advanced Research Projects Activity (IARPA) via Department of Interior/Interior Business Center (DOI/IBC) under 140D0423C0075, by the Amazon Science Hub, by the MIT-Google Program for Computing Innovation, by AMD via the MIT AI Hardware Program, and by a 2025 MIT Office of Research Computing and Data Seed Grant. The views and conclusions contained in this document are those of the authors and should not be interpreted as necessarily representing the official policies, either expressed or implied, of any other entity.

{
\small

\bibliographystyle{unsrtnat} 
\bibliography{main}
}


\appendix

\section{Appendix}
\subsection{Test Set Generation}
\label{subsec:test_set_generation}
Our test set consists of 1,000 videos whose trajectories feature high overlap between the context (i.e., the first 256 frames) and the remaining 768 frames.
We use a near-identical script to generate the training and test sets, but restrict the test set's trajectories to ones that are likely to feature high overlap.
Specifically, for each test video we generate, we first sample 100 complete action sequences.
We then compute these sequences' expected top-down trajectories (assuming movement on a flat plane where there are no collisions with trees, caves, hillsides, etc.).
For each of these expected trajectories, we compute the Chamfer distance between the context's XY points and the remaining XY points.
We then sample uniformly from the 10 sequences with the lowest Chamfer distances to select a single action sequence that is ultimately rendered.
Using this procedure, we render 10,000 test set candidates.
We then choose the 1,000 videos with the highest distance covered by the agent.
This ensures that our test set videos are unlikely to contain trajectories where the agent becomes ``stuck'' by falling into a cave or repeatedly running into a tree.
We observe that this procedure does a reasonable job of producing interesting videos with good context overlap.

We considered using camera frustum overlap~\cite{xiao2025worldmem,song2025generative} in order to measure context overlap, but ultimately decided against doing so because we found that frustum overlap behaves counterintuitively for trajectories (like ours) that feature many loops.
When there are many loops, the context's viewing frustums tend to cover almost every angle, making frustum overlap a meaningless metric.

More broadly, we argue that while it is desirable to ensure good overlap between the context frames and the remaining frames, failing to achieve perfect overlap is not catastrophic.
In cases where the remaining frames have little visual overlap with the context, even a near-perfect generative model can do no better than producing plausible hallucinations.
Consequently, the resulting metrics will primarily reflect generation quality rather than consistency (i.e., in the best case, the metrics will be similar to those computed with respect to random ground-truth frames).
However, because every method is evaluated on the same test set, this effect applies roughly equally across methods: it compresses the absolute values of consistency metrics, effectively making them appear worse, but preserves their relative ordering.
The absolute consistency numbers are therefore harder to interpret in isolation, but cross-method comparisons remain meaningful.
Thus, although we try our best to ensure good overlap, it is not strictly necessary to do so.

\subsection{Implementation and Training Details}
\label{subsec:training_details}

Our models are implemented in PyTorch \cite{paszke2019pytorchimperativestylehighperformance} and use bfloat16 automatic mixed precision (AMP).

\paragraph{Hierarchical Autoencoder Training Setup}
Our hierarchical autoencoder processes each frame in a video independently. It has a ViT-B \cite{dosovitskiy2020image} transformer architecture for both its encoder and decoder. Each input token to both the transformer encoder and decoder gets height and width 2D sin/cos positional embeddings \cite{transformer, kaimingmae}. Each token also has a learned position embedding corresponding to its level in the coarse-to-fine hierarchy. In all our experiments, we set $L=4$. The decoder is further conditioned on the level embedding using AdaLN \cite{peebles2023scalable}. 

Our model operates on images with maximum resolution $(256, 256)$ with a patch size of $(16, 16)$. Each output token has $32$ channels. Our hierarchical autoencoder is trained using the Muon \cite{jordan2024muon} optimizer for linear layers and AdamW \cite{adamw} for the remaining parameters. We use a learning rate of $1\mathrm{e}{-3}$, weight decay of $0.01$, and gradient norm clipping at $1.0$, with a linear warmup over $2{,}000$ steps. The model is trained for $128{,}000$ steps at a per-device batch size of $224$ for $\sim$ 1 day on $8$ H200 GPUs, with an effective batch size of $224 *8 = 1792$.

\paragraph{Generative Model Training Setup}
We instantiate our generative model as a diffusion model \cite{jaschadiffusion, ddpm} with a velocity-field parameterization (v-prediction) \cite{vpred}.
For the noising schedule, we adopt the cosine schedule from Simple Diffusion~\cite{hoogeboom2023simple}, and we empirically find that a shifted variant with shift = 1.0 performs well across all the prediction and upsampling tasks (our schedule is level-independent).
At training time, we randomly drop out all context tokens for a batch element with probability $0.1$ to experiment with classifier-free guidance (CFG) \cite{cfg_paper} at inference time. 

The diffusion model is parameterized using the DiT-B~\cite{peebles2023scalable} architecture, and is trained jointly on latents from all four levels produced by the hierarchical autoencoder.
In the transformer, every token gets a 3D sin/cos position embedding \cite{kaimingmae, transformer} describing its frame, height and width indices.
Each token also gets a learned level embedding. Conditioning on categorical actions (left, right, forward) is provided through a simple learned embedding layer.
Both the level embedding and the diffusion timestep (which is also encoded via sin/cos positional embedding) are passed to the transformer via AdaLN conditioning.

All generative models are trained using the Muon \cite{jordan2024muon} optimizer for linear layers and AdamW \cite{adamw} for the remaining parameters.
We use a learning rate of $1\mathrm{e}{-3}$, weight decay of $0.01$, and gradient norm clipping at $1.0$, with a linear warmup over $2{,}000$ steps.
Each model is trained with a per-device batch size of $48$ on $8$ H200 GPUs, with an effective batch size of $48 *8 = 384$.

\paragraph{Inference Details}
At sampling time we use deterministic DDIM sampling \cite{ddim} with $50$ denoising steps.
We use the same shifted schedule as at training time to decide where to evaluate the 50 denoising steps.
We do not apply CFG to the samples used for evaluation, although we find that sampling quality can be marginally improved by using CFG with guidance scale $\gamma=1.5$.

Examples in our test set contain 1024 total frames.
We always start evaluating the models at a fixed frame (frame 256), with each model having access to a different number of maximum context frames before 256, depending on the particular choices for how each model allocates its context budget.
We then keep rolling out for inference until frame 1024 is generated, appending additional actions where needed.
All comparison metrics are run for the same indices of generated frames $> 256$, comparing to the ground truth.

\paragraph{Model-Specific Details}
Each model is trained with a maximum sequence length $S$.
We list the model-specific implementation details below.
We choose to keep $S$ as close as possible between models to ensure fair comparison.
However, due to details in the nature of the inference strategy, there will be a small discrepancy in the exact values of $S$ for each model.

For the models reported in Figure \ref{fig:baselines_quantitative} and Table \ref{tab:baselines}:

\textbf{Our model (\methodname)}: The model is trained with $S = 3840 $. This $S$ is split between the 3072 tokens reserved for context (which include enough budget for $[3, 12, 48, 192]$ frames at levels 0-3 respectively) and 768 tokens reserved for prediction in any given step. At inference time, the model sees a maximum of $3 + 12 + 48 + 192 = 255$ frames into the past. 

\textbf{FramePack} \cite{zhang2025framepack}: The model is trained with $S = 3840$.
Like for \methodname, 3072 of these tokens are reserved for context budget (enough budget for $[3, 12, 48, 192]$ context frames at levels 0-3 respectively).
The model predicts 3 frames into the future at the finest level, which accounts for the remaining $3 * 256 = 768$ tokens of the sequence length.
At inference time, the model sees a maximum of $3 + 12 + 48 + 192 = 255$ frames into the past. 

\textbf{Full-Resolution Autoregressive Rollout}: The model is trained with $S = 3584$, which is enough for 14 frames at our finest token level.
Following prior works \cite{wang2026matrixgame30realtimestreaming, chen2024diffusion, chen2023videocrafter1}, we allocate half the budget to context frames and half the budget to predicted frames.
At inference time, the model sees a maximum of 7 frames into the past. 

Our main models are trained for $192{,}000$ steps, which takes $\sim$ 2 days on 8 H200 GPUs.
For ablations and comparisons reported in Figure \ref{fig:ablation_cascaded} and Figure \ref{fig:framepack_quantitative}, we opted to train smaller models with a third of the sequence length budget $S$.
These smaller models are trained for $256{,}000$ steps on 8 H200 GPUs with a batch size of $48$ per device (for an effective batch size of $8*48=384$).
This takes $\sim$ 1 day per model.

\subsection{Sampling Speed}
\label{subsec:sampling_speed}
At inference time, the total number of sampling steps differs between the different methods.
We analyze the number of function evaluations needed to run each method and provide the wall clock time needed to run batched inference for $8$ samples with $50$ DDIM \cite{ddim} steps per rollout step.

Full-resolution autoregressive rollout looks at context and predicts tokens only at the highest resolution.
For some fixed constant $k_b$, which depends on context and prediction length, it takes $F /k_b = O(F)$ steps to generate $F$ frames.
For the sampling specification described above, it takes approximately 11 minutes to sample a batch of 8 videos (of 768 frames each).

FramePack \cite{zhang2025framepack} also looks at context and predicts a small number of frames at the finest resolution. As such, for some fixed constant $k_p$ which depends on the context and prediction length, it too takes $F / k_p = O(F)$ steps.
For the sampling specification described above, it takes about 30 minutes to sample a batch of 8 videos (of 768 frames each).

Our method predicts at the coarsest resolution first, and then follows an upsampling strategy level-by-level to predict finer tokens.
As a result, it must not only roll out for $O(F)$ steps at the finest level, but also roll out at the coarser levels.
Fortunately, the number of rollout steps taken at the coarser levels decreases exponentially with every level.
Thus, for our model, for some constant $k_m$, it takes $1/{k_m} (F + F/4 + F/16 + F/64 + \ldots) = O(F)$ steps to generate $F$ frames.
For the sampling specification described above, it takes about 30 minutes to sample a batch of 8 videos (of 768 frames each).
Note that our model requires about 33\% more sampling steps than the equivalent FramePack model, so an optimized FramePack model should require about 25\% less time than ours.

\begin{figure*}[p]
\vspace{-20pt}
\noindent\hspace*{-0.2639in}%
\makebox[0pt][l]{%
\includegraphics{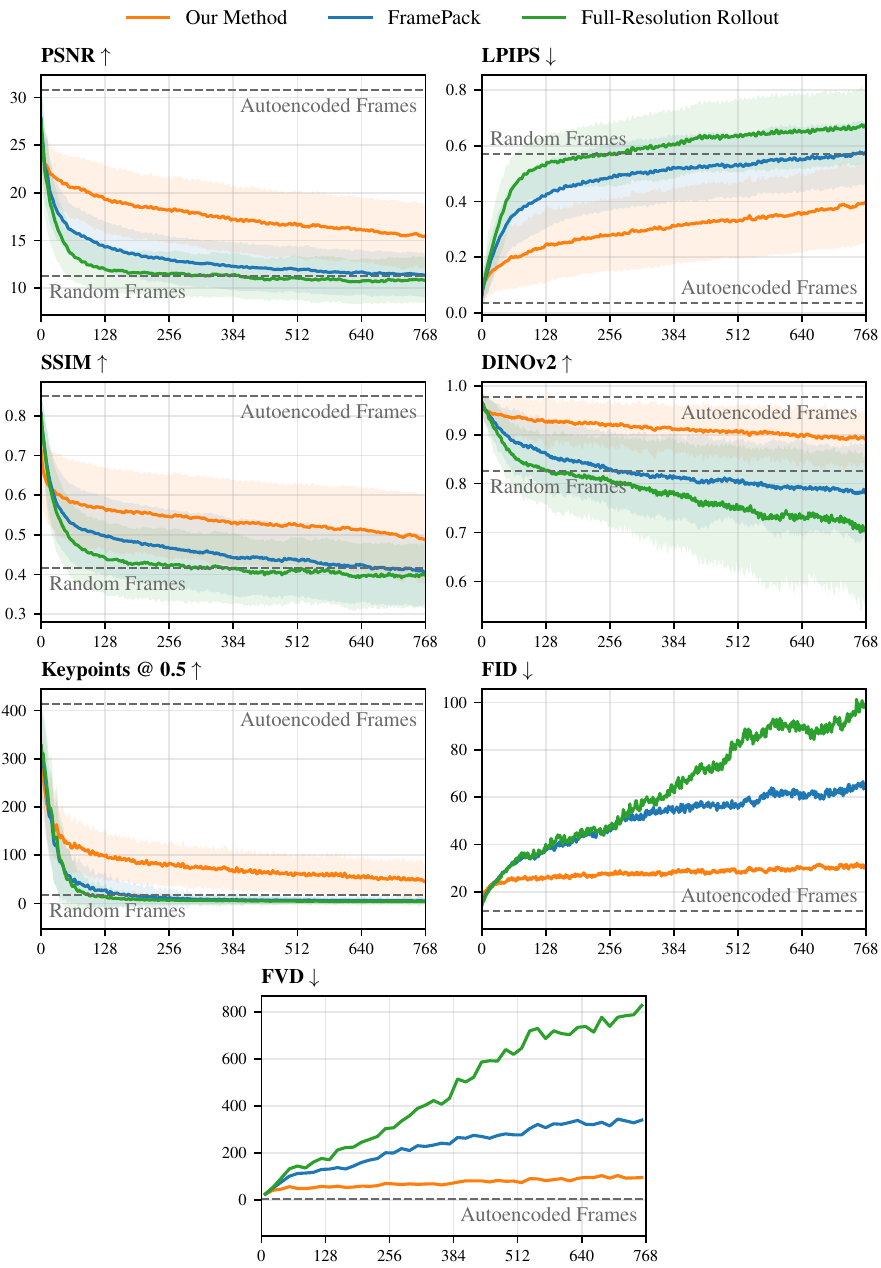}%
}%
\caption{
\textbf{Full metrics:}
All metrics plotted for our method and the baselines.
Our method not only produces better consistency (as measured by PSNR, LPIPS, SSIM, DINOv2 class token cosine similarity, and the number of keypoints with $\geq0.5$ confidence detected by LightGlue), but also produces significantly better per-frame quality as measured by FID and FVD, with less exposure bias.
} 
\label{fig:baselines_quantitative_extended}
\end{figure*}
\begin{figure*}[p]
\noindent\hspace*{-0.0118in}%
\makebox[0pt][l]{%
\includegraphics{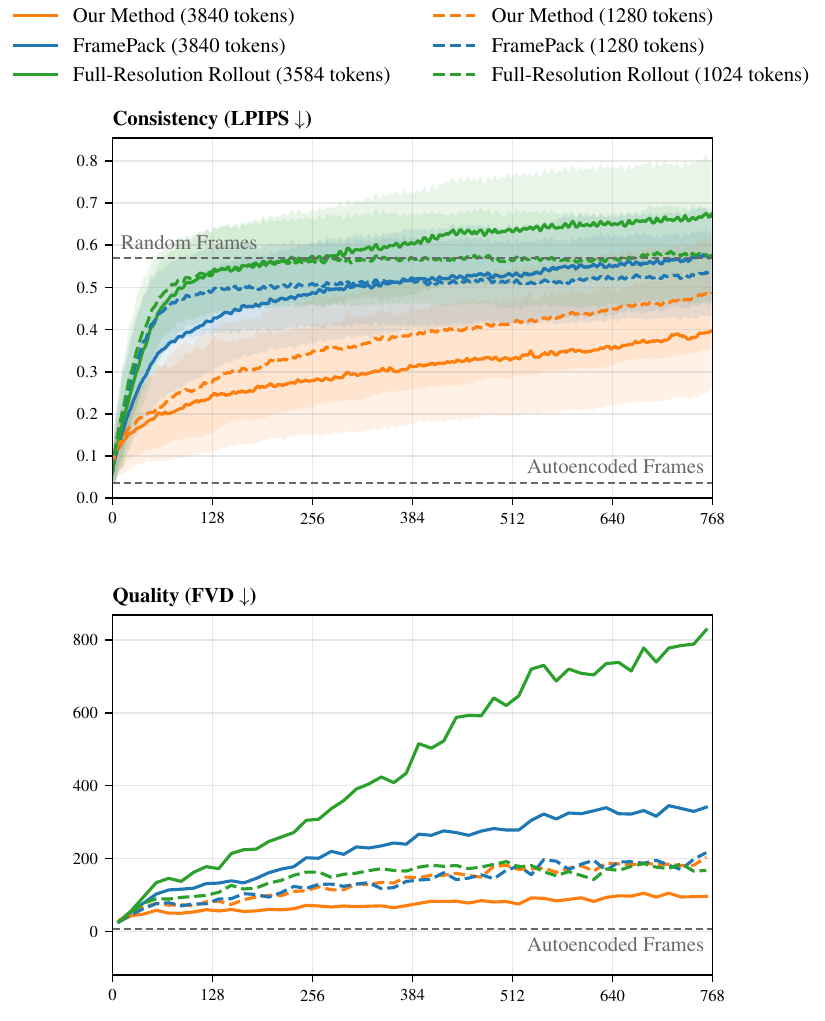}%
}%
\caption{
\textbf{Scaling properties:}
A comparison of our method and the baselines at two distinct scales.
Given 3840 tokens, both our method and FramePack hold 255 frames of context.
Meanwhile, given 3584 tokens, the autoregressive high-resolution rollout model holds 7 frames of context.
Given 1280 tokens, both our method and FramePack hold 85 frames of context; the high-resolution rollout model holds 2 frames of context.
The models with longer sequence length were trained for $192{,}000$ steps, while the ones with shorter sequence length were trained for $256{,}000$ steps.
Longer sequence length improves consistency for both our model and FramePack.
However, it causes the high-resolution rollout model to become unstable, and so its consistency falls below the consistency of random ground-truth frames.
With a lower sequence length, all models have roughly the same per-frame quality.
However, increasing sequence length appears to increase exposure bias (as indicated by increasing FVD) for the baselines while \emph{reducing} it for our model.
This suggests that our model may have more favorable scaling properties than the baselines.
} 
\label{fig:baselines_quantitative_extended}
\end{figure*}
\begin{figure*}[p]
\centering

\makebox[\textwidth][c]{
  \hspace{-18pt}
  \includegraphics[width=436pt]{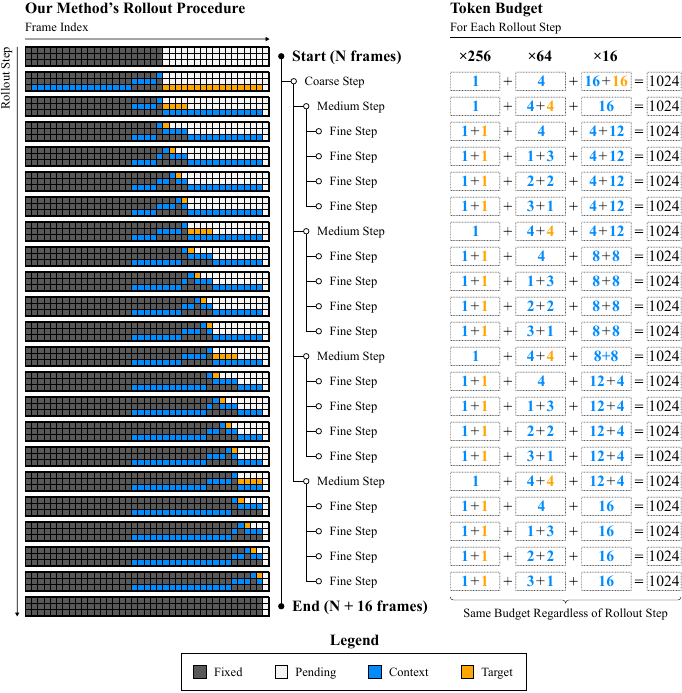}
}
\caption{
\textbf{Our model's rollout procedure:}
On the left, see our model's full rollout sequence in the case of 3 hierarchy levels.
We show the initial state, containing 22 frames; the final state, containing 38 frames; and 21 rollout steps, each of which is represented by a 3-row grid.
During each rollout step, the model sees a mixture of \context{context} and \target{target} tokens across different hierarchy levels; it does not see the \fixed{fixed} (already generated) or \pending{pending} (not yet generated) tokens.
In each step, the rows represent hierarchy levels---fine (top), medium (middle), and coarse (bottom)---with 256, 64, and 16 tokens per frame respectively.
Depending on the step, the model generates long sequences of coarse frames, medium-length sequences of medium-compression frames, or short sequences of fine frames.
The rollout procedure is repeated to generate longer videos.
\underline{Right side:}
During each rollout step, the transformer sees the exact same number of tokens.
Note that while this figure shows the case of 3 levels and 256 tokens at the finest level, our approach scales to arbitrary numbers of levels and different per-image token counts.
Given an increased sequence length budget, one can extend the context length by either (1) multiplying the number of frames at each level by some integer value or (2) increasing the number of levels.
For a more intuitive view of the algorithm, we highly encourage readers to watch the animated version of the rollout sequence on our project website.
} 
\label{fig:diffusion_sampling_long}
\end{figure*}
\begin{figure*}[p]
\centering
\hspace{-8.5pt}
\includegraphics[width=368pt]{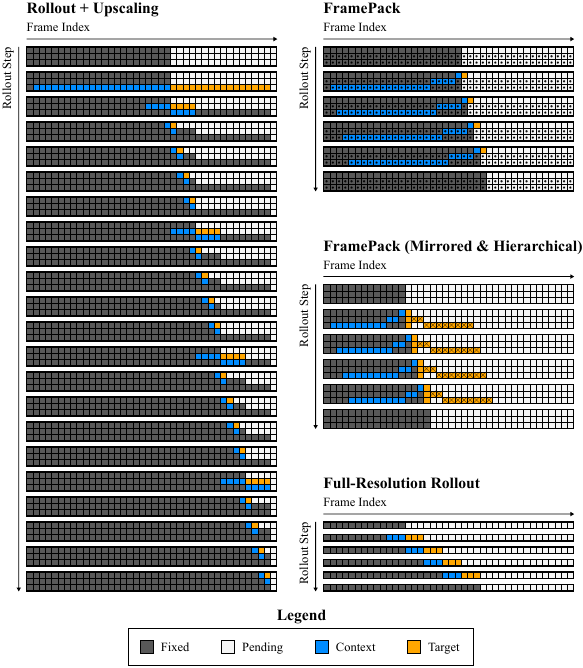}
\caption{
\textbf{Other methods:}
Visualizations of the rollout procedures for the other methods mentioned in the paper.
The \underline{rollout + upscaling} model rolls out at the coarsest level, then combines rollout and upscaling for all subsequent levels.
\underline{FramePack} uses varying patchification, denoted via asterisks, to control the number of tokens per frame.
It always predicts frames at their full resolution.
\underline{FramePack (Mirrored \& Hierarchical)} is a variant of FramePack that uses hierarchical latents.
It is trained to predict coarse frames far into the future.
At sampling time, the coarse predictions are discarded, as indicated by the crosses drawn through them.
In \underline{full-resolution rollout}, there is only a single latent resolution; frames are rolled out at this resolution.
}
\label{fig:other_models}
\end{figure*}
\clearpage


\newpage

\end{document}